\documentclass[oribibl]{llncs}
\usepackage{makeidx}  

\usepackage{graphicx}
\usepackage{amssymb,amsmath}
\usepackage{hyperref}
\usepackage{booktabs}
\usepackage{caption}

\usepackage{todonotes}
\usepackage{comment}

\usepackage{bm}

\newlength{\tab}
\usepackage{array}

\begin{document}
\pagestyle{plain}
\frontmatter          
\mainmatter              
\title{Roto-Translation Covariant Convolutional Networks for Medical Image Analysis}

\titlerunning{Roto-Translation Covariant Convolutional Networks for Medical Image Analysis}  
%
\author{Erik J Bekkers$^{*,1}$, Maxime W Lafarge$^{*,2}$, Mitko Veta$^2$, Koen AJ Eppenhof$^2$, Josien PW Pluim$^2$ and Remco Duits$^1$}
\index{Bekkers, Erik J}
\index{Lafarge, Maxime W}
\index{Veta, Mitko}
\index{Eppenhof, Koen AJ}
\index{Pluim, Josien}
\index{Duits, Remco}

\authorrunning{EJ Bekkers, MW Lafarge, Koen AJ Eppenhof, M Veta, JPw Pluim and R Duits} 
%
\tocauthor{Erik J Bekkers, Maxime W Lafarge, Koen AJ Eppenhof, Mitko Veta, Josien Pw Pluim and Remco Duits}
%
\institute{
Eindhoven University of Technology, $^1$Department of Mathematics and Computer Science and $^2$Department of Biomedical Engineering, Eindhoven, The Netherlands\\%
$^*$Joint main authors - \email{e.j.bekkers@tue.nl}, \email{m.w.lafarge@tue.nl}}

\maketitle              

\hyphenation{e--stab--lished}
\begin{abstract}
We propose a framework for rotation and translation covariant deep learning using $SE(2)$ group convolutions. The group product of the special Euclidean motion group $SE(2)$ describes how a concatenation of two roto-translations results in a net roto-translation. We encode this geometric structure into convolutional neural networks (CNNs) via $SE(2)$ group convolutional layers, which fit into the standard 2D CNN framework, and which allow to generically deal with rotated input samples without the need for data augmentation. 

We introduce three layers: a \emph{lifting layer} which lifts a 2D (vector valued) image to an $SE(2)$-image, i.e., 3D (vector valued) data whose domain is $SE(2)$; a \emph{group convolution layer} from and to an $SE(2)$-image; and a \emph{projection layer} from an $SE(2)$-image to a 2D image.
The lifting and group convolution layers are $SE(2)$ \emph{covariant} (the output roto-translates with the input).
The final projection layer, a maximum intensity projection over rotations, makes the full CNN rotation \emph{invariant}.

We show with three different problems in histopathology, retinal imaging, and electron microscopy that with the proposed group CNNs, state-of-the-art performance can be achieved, without the need for data augmentation by rotation and with increased performance compared to standard CNNs that do rely on augmentation.
\end{abstract}
\section{Introduction}

In this work we generalize $\mathbb{R}^2$ convolutional neural networks (CNNs) to $SE(2)$ group CNNs (G-CNNs) in which the data lives on position orientation space, and in which the convolution layers are defined in terms of representations of the special Euclidean motion group $SE(2)$. In essence this means that we replace the convolutions (with translations of a kernel) by $SE(2)$ group convolutions (with roto-translations of a kernel). The advantage of the proposed approach compared to standard $\mathbb{R}^2$ CNNs is that rotation covariance is encoded in the network design and does not have to be learned by the convolution kernels. E.g., a feature that may appear in the data under several orientations does not have to be learned for each orientation, but only once. As a result, there is no need for data augmentation by rotation and the kernel weights (that no longer need to learn rotation covariance) become available to increase the CNNs expressive capacity. Moreover, the proposed group convolution layers are compatible with standard CNN modules, allowing for easy integration in popular CNN designs.

A main objective of medical image analysis is to develop models that are invariant to the shape and appearance variability of the structures of interest, including their arbitrary orientations. Rotation-invariance is a desired property, which our G-CNN framework generically deals with. We show state-of-the-art results with improvement over standard 2D CNNs on three different medical imaging tasks: mitosis detection in histopathology images, vessel segmentation in retinal images and cell boundary segmentation in electron microscopy (EM).

\subsection{Related work}
In relation to other approaches that incorporate rotation invariance/covariance in the network design, such as harmonic networks \cite{worrall2017harmonic}, local transformation invariance learning \cite{sohn2012learning}, deep symmetry nets \cite{gens2014deep}, scattering CNNs \cite{sifre2013rotation,oyallon2013generic}, and warped convolutions \cite{henriques2017warped}, the group convolution approaches \cite{cohen2016group,dieleman2016exploiting,weiler2017learning,sifre2013rotation,oyallon2013generic,bekkers2018template} most naturally extend the standard CNNs by simply replacing the convolution operators.

In the work by Cohen \& Welling \cite{cohen2016group} a comprehensive theoretical framework for G-CNNs is developed for discrete groups whose transformations stay on the pixel grid. In particular their focus was on the wall-paper groups $p4$ (group of translations + $90^\circ$ rotations), for which a G-CNN approach was also developed by Dieleman et al. \cite{dieleman2016exploiting}, and p4m ($p4$ + reflections). In their work it was convincingly demonstrated that including such symmetries, by replacing standard convolutions by group convolutions, substantially increases the network's performance without increasing the number of network variables. Although their theoretical G-CNN framework \cite{cohen2016group} holds for more general groups, their actual application scope was limited to discrete groups that stay on the pixel grid. In this paper, we are not restricted to such groups, but include efficient bi-linear interpolation that allows us to employ the full structure of the continuous roto-translation group $SE(2)$, which we can discretize to the sub-group $SE(2,N)$, with $N$ rotations. Special cases of our framework are standard 2D CNNs when $N=1$ and the $p4$ G-CNNs as proposed in \cite{cohen2016group,dieleman2016exploiting} when $N=4$.

In very recent work, Weiler et al. \cite{weiler2017learning} describe a different approach to $SE(2)$ G-CNNs. Instead of relying on interpolation they used 2D complex-valued steerable kernels, which has the advantage that kernel rotations are exact. A disadvantage is, however, that these kernels are constrained to a specific combination of complex valued basis functions. With our interpolation approach, kernel rotation simply appears in the CNN architecture as a (sparse) matrix-vector multiplication, that maps a set of base weights to a full set of rotated kernels.

In work by Mallat, Oyallon, and Sifre \cite{oyallon2013generic,sifre2013rotation} roto-translation invariant deep networks are formulated in the context of scattering theory. Their design involves a concatenation of separable group convolutions with hand-crafted (but well underpinned) filters, followed by the modulus as activation function. Learning takes place via support vector machines on the generated $SE(2)$ invariant descriptors. In our approach, the filters are learned without restrictions, the convolutions do not have to be separable, and we here use the common ReLU activation function.

In work by Bekkers et al. \cite{bekkers2018template}, an effective template matching method was proposed using group correlations in orientation scores, which are $SE(2)$ images obtained from a 2D image via lifting convolutions with a specific choice of kernel \cite{Duits2007a}. The $SE(2)$ templates were put in a B-spline basis (allowing for exact kernel rotations) and optimized via logistic regression. Their architecture fits within our framework as a single channel G-CNN of depth 2 with a fixed lifting kernel.

\section{$SE(2)$ convolutional neural networks}

\subsection{Group theoretical preliminaries}
\noindent
\textbf{The Lie group $\bm{SE(2)}$}: 
The group $SE(2) = \mathbb{R}^2 \rtimes SO(2)$ is the semi-direct product of the group of planar translations $\mathbb{R}^2$ and rotations $SO(2)$, and its group product is given by
\begin{equation}
g \cdot g' = ( \mathbf{x},\mathbf{R}_\theta ) \cdot ( \mathbf{x}', \mathbf{R}_{\theta'} )
= ( \mathbf{R}_\theta \mathbf{x}' + \mathbf{x}, \mathbf{R}_{\theta + \theta'}),\\
\end{equation}
with group elements $g = (\mathbf{x},\theta),g' = (\mathbf{x}',\theta') \in SE(2)$, with translations $\mathbf{x},\mathbf{x}'$ and planar rotations by $\theta,\theta'$. 
The group acts on the space of positions and orientations $\mathbb{R}^2 \times S^1$ via
$
g \cdot (\mathbf{x}',\theta') = ( \mathbf{R}_\theta \mathbf{x}' + \mathbf{x}, \theta +\theta').
$
Since $( \mathbf{x} , \mathbf{R}_\theta ) \cdot ( \mathbf{0}, 0 ) = ( \mathbf{x} , \theta )$, we can identify the group $SE(2)$ with the space of positions and orientations $\mathbb{R}^2 \times S^1$. As such we will often write $g=(\mathbf{x},\theta)$, instead of $(\mathbf{x},\mathbf{R}_\theta)$. Note that $g^{-1} = ( -\mathbf{R}_{\theta}^{-1} \mathbf{x},-\theta)$ since $g \cdot g^{-1} = g^{-1} \cdot g = (\mathbf{0},0)$.

\begin{figure}[h]
\includegraphics[width=\textwidth]{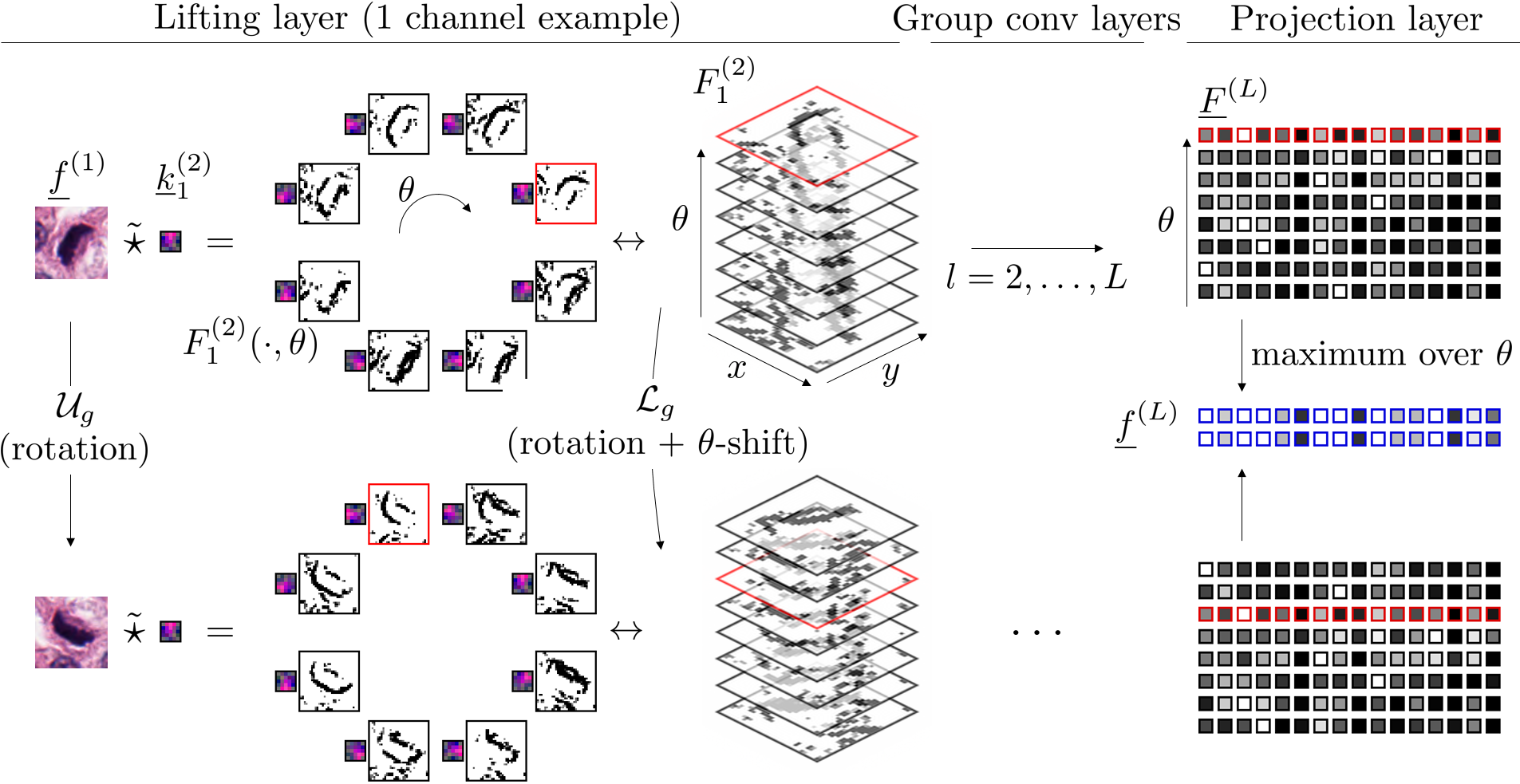}
\centering
\caption{Rotation co- and invariance. Top row: the activations after the lifting convolutions with a single kernel $\underline{k}_1^{(2)}$, stacked together it yields an $SE(2)$ image $F_1^{(2)}$ (cf. Eq.~(\ref{eq:liftingLayer})). The projection layer at the end of the pipeline gives a rotation invariant feature vector. Bottom row: the same figures with a rotated input.}
\label{fig:illustration}
\end{figure}

\noindent
\textbf{Group representations}: 
The structure of the group can be mapped to other mathematical objects (such as 2D images) via representations. The left-regular $SE(2)$ representation on 2D images $f\in \mathbb{L}_2(\mathbb{R}^2)$ is given by
\begin{equation}
\label{eq:leftregreprSE2on2D}
(\mathcal{U}_g f) (\mathbf{x}') = f(\mathbf{R}_\theta^{-1} (\mathbf{x}' - \mathbf{x})),
\end{equation}
with $g = (\mathbf{x},\theta) \in SE(2), \; \mathbf{x}' \in \mathbb{R}^2$. It corresponds to a roto-translation of the image. The left-regular representation on functions $F\in \mathbb{L}_2(SE(2))$ on $SE(2)$, which we refer to as $SE(2)$-images, is given by
\begin{equation}
\label{eq:leftregreprSE2}
(\mathcal{L}_g  F) (g')= F(g^{-1} \cdot g') = F(\mathbf{R}_\theta^{-1} (\mathbf{x}' - \mathbf{x}), \theta' - \theta),
\end{equation}
with $g=(\mathbf{x},\theta),g'=(\mathbf{x}',\theta') \in SE(2)$. It is a shift-twist (rotation + $\theta$-shift) of $F$, see e.g. Fig.~\ref{fig:illustration}. Next we define the G-CNN layers in terms of these representations.

\subsection{The $SE(2)$ group convolution layers}
In CNNs one can take a convolution or a cross-correlation viewpoint and since these operators simply relate via a kernel reflection, the terminology is often used interchangeably. We take the second viewpoint, our G-CNNs are implemented using cross-correlations. On $\mathbb{R}^2$ we define cross-correlation via inner products of translated kernels:
\begin{equation}
(k \star_{\mathbb{R}^2} f)(\mathbf{x})
:= (\mathcal{T}_\mathbf{x} k, f)_{\mathbb{L}_2(\mathbb{R}^2)} 
:= \int_{\mathbb{R}^2} k(\mathbf{x}' - \mathbf{x}) f(\mathbf{x}') {\rm d}\mathbf{x}',
\end{equation}
with $\mathcal{T}_\mathbf{x}$ the translation operator, the left-regular representation of the translation group $(\mathbb{R}^2,+)$. In the $SE(2)$ lifting layer we now simply replace translations of $k$ by roto-translations via the $SE(2)$ representation $\mathcal{U}_g$ defined in Eq.~(\ref{eq:leftregreprSE2on2D}).

\noindent
\textbf{The $\bm{SE(2)}$ lifting layer}: 
Let $\underline{f},\underline{k}:\mathbb{R}^2\rightarrow \mathbb{R}^{N_c}$ be a vector valued 2D image and kernel (with $N_c$ channels), with $\underline{f} = (f_{1},\dots,f_{N_c})$ and $\underline{k} = (k_{1},\dots,k_{N_c})$, then the group lifting correlations for vector valued images are defined by
\begin{equation}
(\underline{k}\,\tilde{\star} \underline{f})(g) 
:= \sum\limits_{c=1}^{N_c} ( \mathcal{U}_g  k_c, f_c )_{\mathbb{L}_2(\mathbb{R}^2)}
= \sum\limits_{c=1}^{N_c} \int_{\mathbb{R}^2} k_c(\mathbf{R}_\theta^{-1} (\mathbf{y} - \mathbf{x})) f_c(\mathbf{y}) {\rm d}\mathbf{y}.
\end{equation}
These correlations \emph{lift} 2D image data to data that lives on the 3D position orientation space $\mathbb{R}^2\times S^1 \equiv SE(2)$. The \emph{lifting layer} that maps from a vector image $\underline{f}^{(l-1)}:\mathbb{R}^2 \rightarrow \mathbb{R}^{N_{l-1}}$, with $N_{l-1}$ channels at layer $l-1$, to an $SE(2)$ vector image $\underline{F}^{(l)}:SE(2)\rightarrow\mathbb{R}^{N_l}$ using a set of $N_l$ kernels $\mathbf{k}^{(l)} := (\underline{k}_1^{(l)},\dots,\underline{k}_{N_l}^{(l)})$, each with $N_{l-1}$ channels, is then defined by
\begin{equation}\label{eq:liftingLayer}
\underline{F}^{(l)} = 
\mathbf{k}^{(l)} \tilde{\star} \underline{f}^{(l-1)} := 
\left( \;\;
\underline{k}_{1}^{(l)} \tilde{\star} \underline{f}^{(l-1)} \;\; ,
\;\;\dots\; , \;\;
\underline{k}_{N_{l}}^{(l)} \tilde{\star} \underline{f}^{(l-1)} 
\;\;\right).
\end{equation}

\noindent
\textbf{The $\bm{SE(2)}$ group convolution layer}: 
Let $\underline{F},\underline{K}:SE(2)\rightarrow \mathbb{R}^{N_c}$ be a vector valued $SE(2)$ image and kernel, with $\underline{F} = (F_{1},\dots,F_{N_c})$ and $\underline{K} = (K_{1},\dots,K_{N_c})$, then the group correlations are defined as
\begin{equation}
(\underline{K} \star \underline{F})(g) 
:= \sum\limits_{c=1}^{N_c} ( \mathcal{L}_g K_c, F_c )_{\mathbb{L}_2(SE(2))}
= \sum\limits_{c=1}^{N_c} \int_{SE(2)} K_c(g^{-1} \cdot h) F_c(h) {\rm d}h,
\end{equation}
with $(K, F)_{\mathbb{L}_2(SE(2))} := \int_{SE(2)} K(h) F(h) {\rm d}h$, the inner product on $\mathbb{L}_2(SE(2))$. 
A set of $SE(2)$ kernels $\mathbf{K}^{(l)} := (\underline{K}_1^{(l)},\dots,\underline{K}_{N_l}^{(l)})$ defines a {group convolution} layer, mapping from 
$\underline{F}^{(l-1)}$ with $N_{(l-1)}$ channels to $\underline{F}^{(l)}$ with $N_{(l)}$ channels, via 
\begin{equation}\label{eq:groupConvLayer}
\underline{F}^{(l)} \!= \!
\mathbf{K}^{(l)} {\star} \underline{F}^{(l-1)} \!:= \!
\left(\;\; 
\underline{K}_{1}^{(l)} {\star} \underline{F}^{(l-1)} \;\;,\;\;
\dots \;\;,\;\;
\underline{K}_{N_{l}}^{(l)} {\star} \underline{F}^{(l-1)} 
\;\;\right).
\end{equation}

\noindent
\textbf{The projection layer}: Projects a multi-channel $SE(2)$ image back to $\mathbb{R}^2$ via
\begin{equation}\label{eq:projectionLayer}
\underline{f}^{(l)}(\mathbf{x}) \!= \underset{\theta \in [0,2\pi]}{\operatorname{max}} \; \underline{F}^{(l)}(\mathbf{x},\theta).
\end{equation}

\subsection{Discretization and network architecture}

\textbf{Discretization, kernel sizes and rotation}: 
Discretized 2D images are supported on a bounded subset of $\mathbb{Z}^2 \subset \mathbb{R}^2$ and the kernels live on a spatially rectangular grid of size $n\times n$ in $\mathbb{Z}^2$, with $n$ the kernel size. We discretize the Lie group $SE(2,N):=\mathbb{Z}^2 \rtimes SO(2,N)$, with the space of 2D rotations in $SO(2)$ sampled with $N$ rotation angles $\theta_i=\frac{2\pi}{N}i$, with $i=0,\dots,N-1$. The discrete lifting kernels $\mathbf{k}^{(l)}$ at layer $l$, mapping from a 2D image with $N_{l-1}$ input channels to an $SE(2,N)$ image with $N_{l}$ channels, thus have a shape of $n\times n \times N_{l-1} \times {N_{l}}$. The $SE(2,N)$ kernels $\mathbf{K}^{(l)}$ have a shape of $n \times n \times N \times N_{l-1} \times N_{l}$. A complete set of rotations of kernels $\mathbf{k}^{(l)}$ or $\mathbf{K}^{(l)}$ can be constructed with a single matrix multiplication from a vector that contains the shared kernel weights. This matrix is sparse and encodes bi-linear interpolation and kernel rotation.

\definecolor{grayish}{rgb}{0.9,0.2,0.2}
\begin{table}[h]
\caption{$SE(2,N)$ chain settings for different orientation samplings $N$. 
}
\label{tab:resultsAndWeights}
\centering
\scriptsize
\begin{tabular}{ l | >{\arraybackslash}p{1.95cm}  >{\arraybackslash}p{1.95cm}  >{\arraybackslash}p{1.95cm}  >{\arraybackslash}p{1.95cm}  >{\arraybackslash}p{2.1cm} }
    \toprule
$N$ (Group) & 1 ($\mathbb{Z}^2$) & 2 ($\mathbb{Z}^2 \times p2$) & 4 ($\mathbb{Z}^2 \times p4$) & 8 ($SE(2,8)$) & 16 ($SE(2,16)$) \\\hline\hline
\multicolumn{6}{l}{\hspace{0cm}\color{grayish}\textbf{Layer 1} - lifting with Eq.~(\ref{eq:liftingLayer}), $n = 5$} \\ 
\;\;$N_1$ {\color{black}($\#_w$)} & 16 {\color{black}(1040)} & 13 {\color{black}(845)} & 10 {\color{black}(650)} & 8 {\color{black}(520)} & 6 {\color{black}(390)} \\ 
\multicolumn{6}{l}{\hspace{0cm}\color{grayish} \textbf{Layer 2,3,4} - group conv. with Eq.~(\ref{eq:groupConvLayer}), $n = 5$} \\ 
\;\;$N_2$ {\color{black}($\#_w$)} & 16 {\color{black}(5408)} & 13 {\color{black}(7124)} & 10 {\color{black}(8420)} & 8 {\color{black}(10768)} & 6 {\color{black}(12108)} \\ 
\;\;$N_3$ {\color{black}($\#_w$)} & 16 {\color{black}(5408)} & 13 {\color{black}(7124)} & 10 {\color{black}(8420)} & 8 {\color{black}(10768)} & 6 {\color{black}(12108)} \\ 
\;\;$N_4$ {\color{black}($\#_w$)} & 64 {\color{black}(21632)} & 32 {\color{black}(17536)} & 16 {\color{black}(13472)} & 8 {\color{black}(10768)} & 4 {\color{black}(8072)} \\ 
\multicolumn{6}{l}{\hspace{0cm}\color{grayish}\textbf{Layer 5} - group conv.  with Eq.~(\ref{eq:groupConvLayer}) + projection with Eq.~(\ref{eq:projectionLayer}), $n = 1$} \\
\;\;$N_5$ {\color{black}($\#_w$)} & 16 {\color{black}(1056)} & 16 {\color{black}(1056)} & 16 {\color{black}(1056)} & 16 {\color{black}(1056)} & 16 {\color{black}(1056)} \\ 
\multicolumn{6}{l}{\hspace{0cm}\color{grayish}\textbf{Layer 6} - standard conv. (output) layer, $n = 1$} \\
\;\;$N_6$ {\color{black}($\#_w$)} & 1 {\color{black}(17)} & 1 {\color{black}(17)} & 1 {\color{black}(17)} & 1 {\color{black}(17)} & 1 {\color{black}(17)} \\ \hline\hline
    Total $\#_w$ & 34561 & 33702 & 32035 & 33897 & 33751 \\
    \bottomrule
\end{tabular}
\end{table}

\section{Experiments and Results}

We consider three different tasks in three different modalities.
In each we consider the $SE(2,N)$ samplings with $N\in\{1,2,4,8,16\}$ to study the effect of the choice of $N$ in the $SE(2,N)$ discretization. See Table.~\ref{tab:resultsAndWeights} for the network settings.
In each experiment the data is augmented at train and test time with transposed versions of the 2D input. For reference we also include transpose plus $90^\circ$ rotation augmentation for the $N=1$ experiment (as in \cite{lafarge2017domain,cirecsan2013mitosis}) in order to be able to show that these are not necessary in our $SE(2,N)$ networks for $N\geq 4$.
Each experiment is repeated 3 times with random initialization and sampling to get a rough estimate of the mean and variance on the performance.
For a fair comparison for different $N$ the overall number of weights is matched. For a fair comparison with the $\mathbb{R}^2$ approach, the number of "2D" activations ($N_l N$) in the last three layers is also matched. Each network optimizes a logistic loss using stochastic gradient descent with momentum using the same settings as in \cite{lafarge2017domain}. Our G-CNN implementations are available at {https://github.com/tueimage/se2cnn}. The results are given in Fig.~\ref{fig:results}, the tasks and metrics are summarized as follows.

\newlength{\figheight}
\setlength{\figheight}{40mm}
\newlength{\figwidth}
\setlength{\figwidth}{34mm}
\begin{figure}[h]
\hspace{2mm}\includegraphics[width=\figwidth]{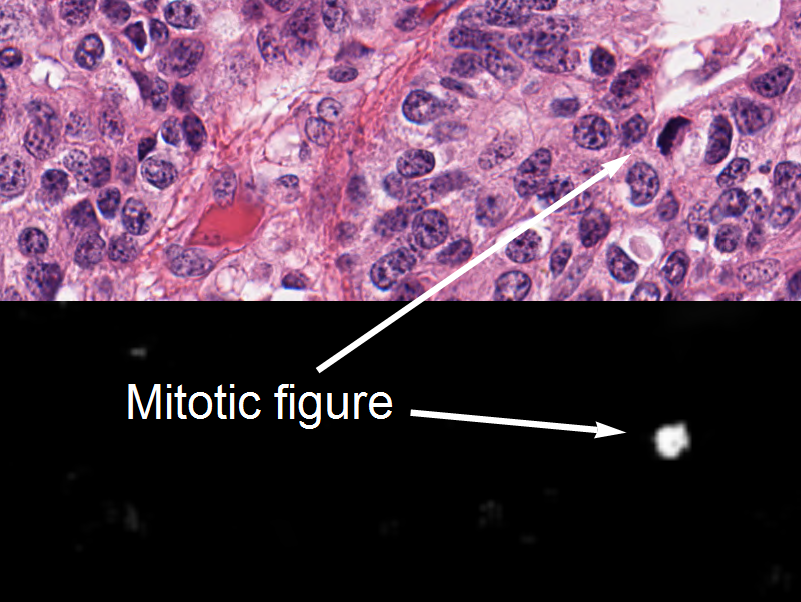}
\hspace{6mm} \includegraphics[width=\figwidth]{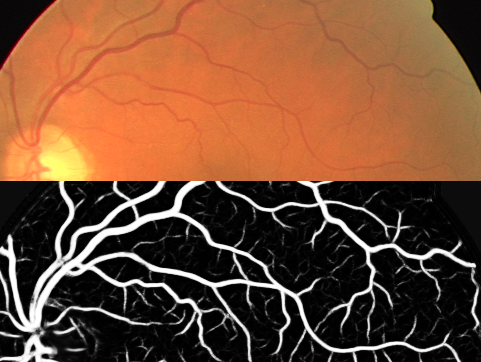}
\hspace{4mm} \includegraphics[width=\figwidth]{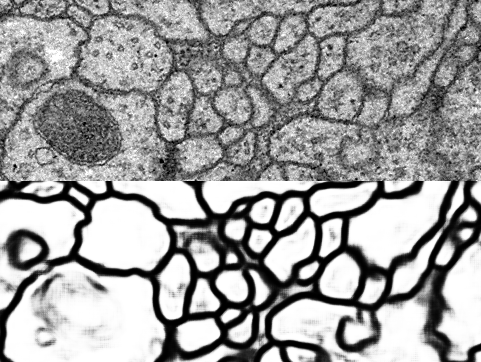}\newline
\includegraphics[height=\figheight]{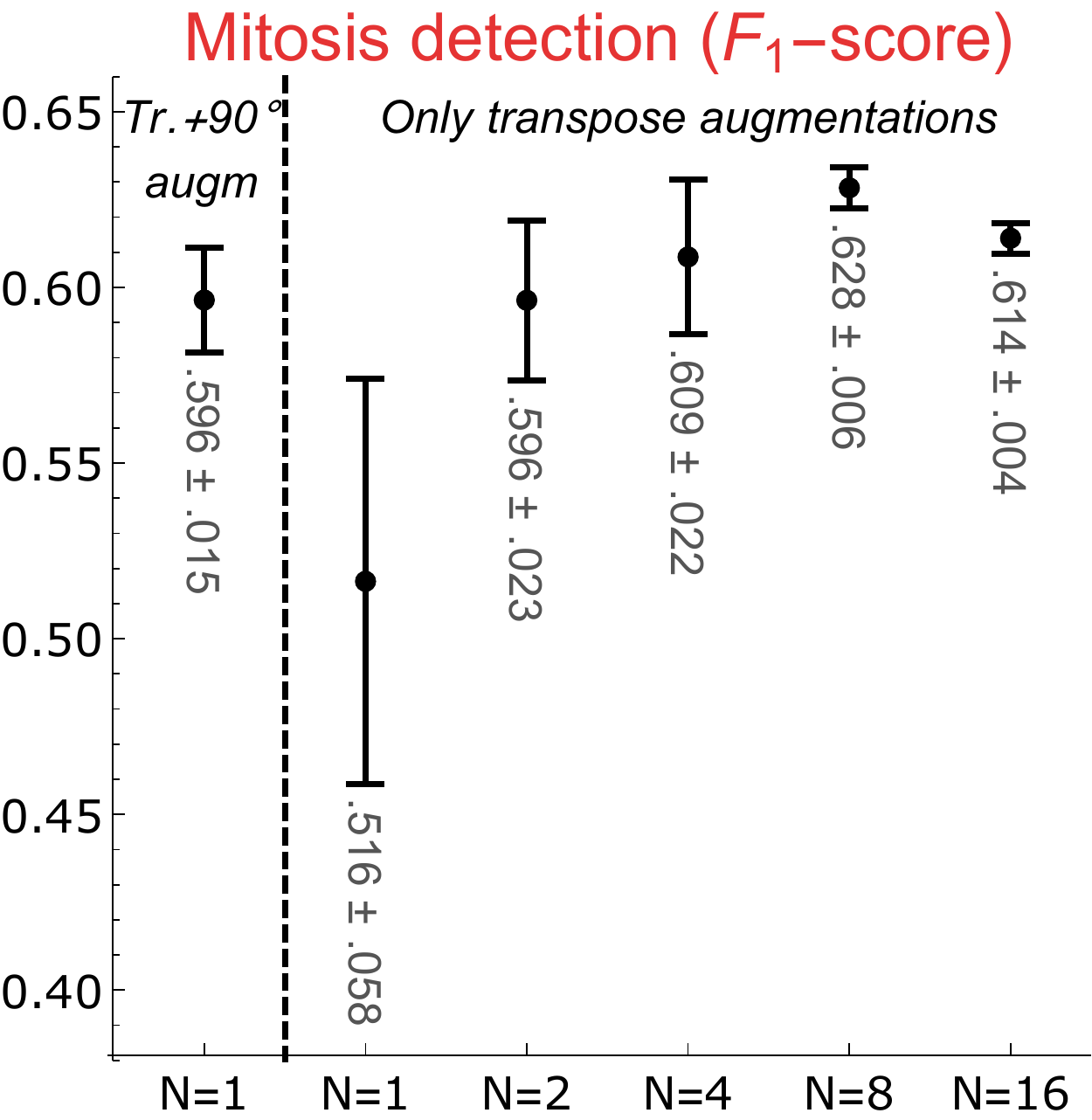}
\includegraphics[height=\figheight]{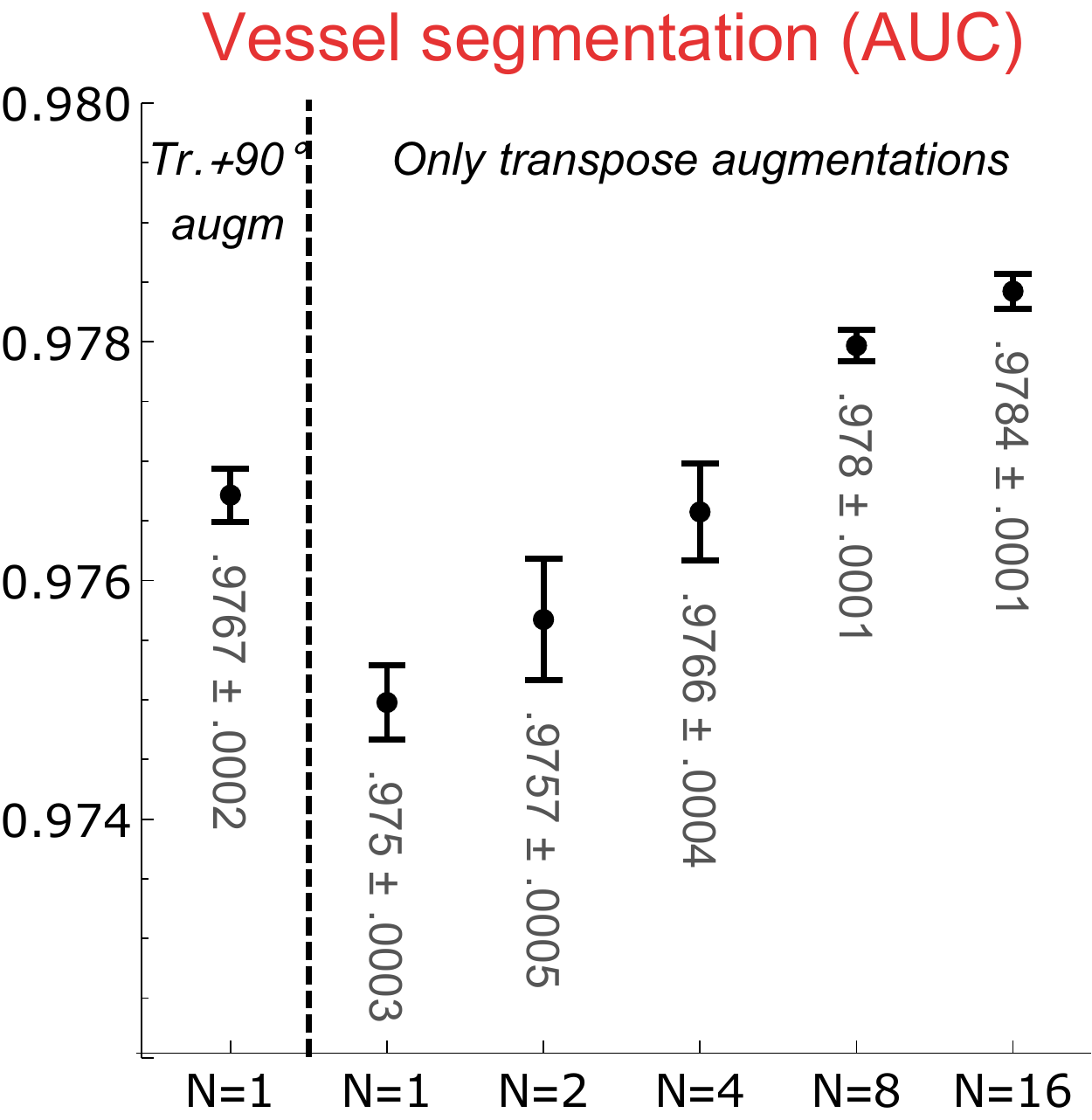}
\includegraphics[height=\figheight]{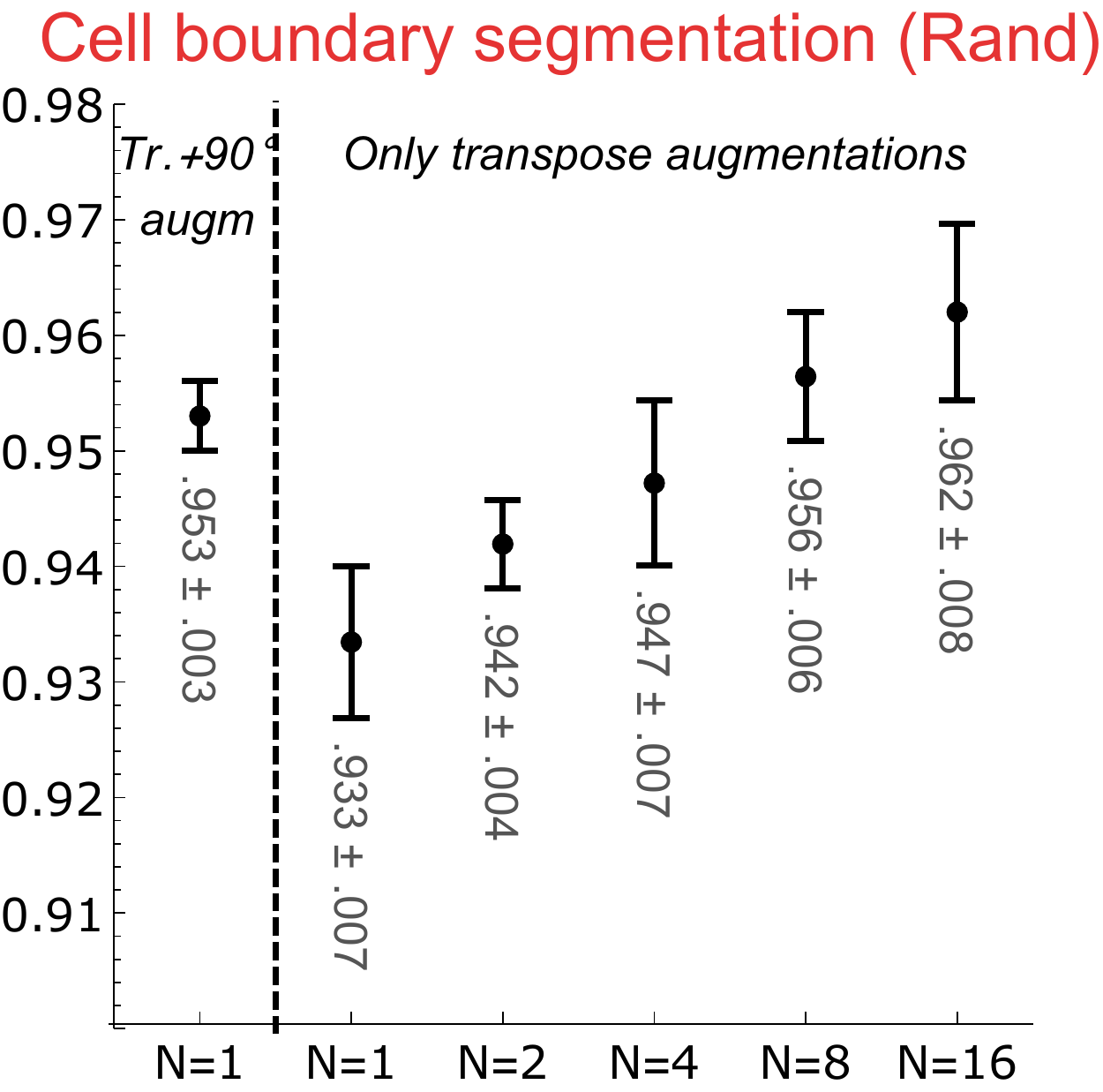}
\centering
\caption{Top row: Crop outs of images of the three tasks with the class probabilities generated by our method. Bottom row: Mean results ($\pm 1$ std. dev.). 
}
\label{fig:results}
\end{figure}

\noindent
\textbf{Histopathology - Mitosis detection:}
The task aims at detecting mitotic figures in hematoxylin-eosin stained slides.
We used the public dataset AMIDA13 \cite{veta2015assessment} that consists of high power field images from 23 breast cancer cases. Eight cases (458 mitoses) were used to train the networks with random batches of $68 \times 68$ image patches, balanced between mitotic and hard negative figures. This receptive field was obtained by means of max-pooling operations in the first three layers.
Sets of candidate detections were generated as in \cite{cirecsan2013mitosis} after selection of an operating point on four validation cases (92 mitoses).
We assessed an F$_{1}$-score for each model based on the 11 test cases (533 mitoses) in the conditions of \cite{veta2015assessment}.

\noindent
\textbf{Retina - Vessel segmentation:}
In this task the blood vessels in the retina are segmented. For validation we use the public DRIVE database \cite{Staal2004Ridge}, which consists of 40 retinal images with manual segmentations. The set is split in a training set (of which we use 16 for training, and 4 for validation) and a test set of also 20 images. The G-CNNs produce a probability for the vessel and background class. Training is done with $10000$ patches ($17 \times 17$) per class per image.
The output probabilities can be thresholded to create a binary segmentation, which can be used to quantify performance in terms of sensitivity and specificity. The area under the receiver operator characteristic (ROC) curve, in short AUC, summarizes these performances into a single value.

\noindent
\textbf{Electron microscopy - Cell boundary segmentation:}
This task consists of segmenting the boundaries of cells that are imaged with EM. We use the data and evaluation system of the ISBI EM segmentation challenge \cite{arganda2015crowdsourcing}. The data consists of 2 volumes (1 train, 1 test), each containing 30 consecutive images from a serial section transmission EM. Both the segmentation and the evaluation is done by treating the volumes as sequences of 2D slices. To increase receptive field size we include max pooling in the first 2 layers. Training is done with $10000$ patches ($48 \times 48$) per class per image.
The main evaluation criterion for the challenge is the Rand score, which measures the similarity between clusterings/connected components \cite{rand1971objective}. The reported Rand score is the maximum score (for several thresholds) computed for the connected components obtained after thinning of the binary cell boundary segmentation, see \cite{arganda2015crowdsourcing} for more details. 

\noindent
\textbf{Results:} In each experiment we see that the performance of the baseline with extra rotation augmentations is reached by the non-augmented G-CNNs for $N\geq 4$, and even is surpassed for $N \geq 8$.
In the first two experiments we also observe that the variance on the output is reduced with increasing $N$.
Our results on the public datasets match or improve upon the state of the art with the following scores: F$_{1}$-score=$0.628 \pm 0.006$ for mitosis detection, AUC = $0.9784 \pm 0.0001$ for vessel segmentation, Rand = $0.962 \pm 0.008$ for cell boundary segmentation.

\section{Discussion and Conclusions}
We showed a consistent improvement of performances across three medical image analysis tasks when using G-CNNs compared to their corresponding CNN baselines.
The reported results are in line with the benchmark of each dataset and the best performances were obtained for an orientation capacity $N \geq 4$, indicating the advantage of learning such rotation-invariant representations. We observed improved stability over the repeated experiments in mitosis detection and vessel segmentation for $N=8$ and $N=16$, suggesting a regularization effect due to the increased weight sharing with increasing $N$.

We conclude that it is beneficiary to include $SE(2)$ group convolution layers in CNN network design, as this avoids the need for rotation augmentation and it improves overall performance. In all three medical imaging problems we achieved state-of-the-art results with the same (basic) network design for each task. Based on these results we expect that our $SE(2)$ layers may lead to a further performance increase when embedded in more complex network designs, such as the popular UNets and ResNets.

{
\noindent
\textbf{Acknowledgements:}
The research leading to these results has received funding from the ERC council under the EC's 7th Framework Programme (FP7/2007--2013) / ERC grant agr. No. 335555.}

%
\bibliographystyle{splncs}
\bibliography{references}

\begin{thebibliography}{10}

\bibitem{worrall2017harmonic}
Worrall, D.E., Garbin, S.J., Turmukhambetov, D., Brostow, G.J.:
\newblock Harmonic networks: Deep translation and rotation equivariance.
\newblock In: CVPR. (2017)  5028--5037

\bibitem{sohn2012learning}
Sohn, K., Lee, H.:
\newblock Learning invariant representations with local transformations.
\newblock In: CVPR, Omnipress (2012)  1339--1346

\bibitem{gens2014deep}
Gens, R., Domingos, P.M.:
\newblock Deep symmetry networks.
\newblock In: Advances in neural information processing systems. (2014)
  2537--2545

\bibitem{sifre2013rotation}
Sifre, L., Mallat, S.:
\newblock Rotation, scaling and deformation invariant scattering for texture
  discrimination.
\newblock In: CVPR, IEEE (2013)  1233--1240

\bibitem{oyallon2013generic}
Oyallon, E., Mallat, S., Sifre, L.:
\newblock Generic deep networks with wavelet scattering.
\newblock arXiv preprint arXiv:1312.5940 (2013)

\bibitem{henriques2017warped}
Henriques, J.F., Vedaldi, A.:
\newblock Warped convolutions: Efficient invariance to spatial transformations.
\newblock In: Int. Conf. on Machine Learning. (2017)  1461--1469

\bibitem{cohen2016group}
Cohen, T., Welling, M.:
\newblock Group equivariant convolutional networks.
\newblock In: Int. Conf. on Machine Learning. (2016)  2990--2999

\bibitem{dieleman2016exploiting}
Dieleman, S., De~Fauw, J., Kavukcuoglu, K.:
\newblock Exploiting cyclic symmetry in convolutional neural networks.
\newblock arXiv preprint arXiv:1602.02660 (2016)

\bibitem{weiler2017learning}
Weiler, M., Hamprecht, F.A., Storath, M.:
\newblock Learning steerable filters for rotation equivariant cnns.
\newblock arXiv preprint arXiv:1711.07289 (2017)

\bibitem{bekkers2018template}
Bekkers, E.J., Loog, M., ter Haar~Romeny, B.M., Duits, R.:
\newblock Template matching via densities on the roto-translation group.
\newblock IEEE tPAMI \textbf{40}(2) (2018)  452--466

\bibitem{Duits2007a}
Duits, R., Felsberg, M., Granlund, G.H., ter Haar~Romeny, B.M.:
\newblock Image analysis and reconstruction using a wavelet transform
  constructed from a reducible representation of the {Euclidean} motion group.
\newblock IJCV \textbf{72}(1) (2007)  79--102

\bibitem{lafarge2017domain}
Lafarge, M.W., Pluim, J.P., Eppenhof, K.A., Moeskops, P., Veta, M.:
\newblock Domain-adversarial neural networks to address the appearance
  variability of histopathology images.
\newblock In: MICCAI-DLMIA 2017.
\newblock Springer (2017)  83--91

\bibitem{cirecsan2013mitosis}
Cire{\c{s}}an, D.C., Giusti, A.,  et~al.:
\newblock Mitosis detection in breast cancer histology images with deep neural
  networks.
\newblock In: MICCAI, Springer (2013)  411--418

\bibitem{veta2015assessment}
Veta, M., {van}~Diest, P., Willems, S.,  et~al.:
\newblock Assessment of algorithms for mitosis detection in breast cancer
  histopathology images.
\newblock MEDIA \textbf{20}(1) (2015)  237--248

\bibitem{Staal2004Ridge}
Staal, J., Abr{\`a}moff, M.D., Niemeijer, M.,  et~al.:
\newblock Ridge-based vessel segmentation in color images of the retina.
\newblock IEEE TMI \textbf{23}(4) (2004)  501--509

\bibitem{arganda2015crowdsourcing}
Arganda-Carreras, I., Turaga, S.C.,  et~al.:
\newblock Crowdsourcing the creation of image segmentation algorithms for
  connectomics.
\newblock Front. in neuroanatomy \textbf{9} (2015)  142

\bibitem{rand1971objective}
Rand, W.M.:
\newblock Objective criteria for the evaluation of clustering methods.
\newblock Journal of the American Statistical association \textbf{66}(336)
  (1971)  846--850

\end{thebibliography}

\end{document}